\documentclass[conference]{IEEEtran}
\IEEEoverridecommandlockouts

\usepackage{cite}
\usepackage{amsmath,amssymb,amsfonts}
\usepackage{algorithmic}
\usepackage{graphicx}
\usepackage{textcomp}
\usepackage{xcolor}
\def\BibTeX{{\rm B\kern-.05em{\sc i\kern-.025em b}\kern-.08em
    T\kern-.1667em\lower.7ex\hbox{E}\kern-.125emX}}

\IEEEoverridecommandlockouts \IEEEpubid{\makebox[\columnwidth]{979-8-3315-3153-9/24/\$31.00~\copyright~2024~IEEE \hfill} \hspace{\columnsep}\makebox[\columnwidth]{ }}

\begin{document}

\title{Comparative Analysis of Deep Learning Approaches for Harmful Brain Activity Detection Using EEG\\
}

\author{
\IEEEauthorblockN{
    Shivraj Singh Bhatti\IEEEauthorrefmark{1}, 
    Aryan Yadav\IEEEauthorrefmark{2}, 
    Mitali Monga\IEEEauthorrefmark{3}, 
    Neeraj Kumar\IEEEauthorrefmark{4}
}
\IEEEauthorblockA{
    \IEEEauthorrefmark{1}\IEEEauthorrefmark{2}\IEEEauthorrefmark{3}\IEEEauthorrefmark{4}
    Department of Computer Science and Engineering, \\
    Thapar Institute of Engineering and Technology, Patiala, Punjab, India \\
    Email: sbhatti\_be18@thapar.edu, ayadav\_be18@thapar.edu, mmonga\_be18@thapar.edu, neeraj.kumar@thapar.edu
}
}

\maketitle

\begin{abstract}
The classification of harmful brain activities, such
as seizures and periodic discharges, play a vital role in
neurocritical care, enabling timely diagnosis and intervention.
Electroencephalography (EEG) provides a non-invasive method
for monitoring brain activity, but the manual interpretation of
EEG signals are time-consuming and rely heavily on expert
judgment. This study presents a comparative analysis of deep
learning architectures, including Convolutional Neural Networks
(CNNs), Vision Transformers (ViTs), and EEGNet, applied to the
classification of harmful brain activities using both raw EEG data
and time-frequency representations generated through Continuous Wavelet Transform (CWT). We evaluate the performance of
these models use multimodal data representations, including
high-resolution spectrograms and waveform data, and introduce
a multi-stage training strategy to improve model robustness.
Our results show that training strategies, data preprocessing,
and augmentation techniques are as critical to model success as
architecture choice, with multi-stage TinyViT and EfficientNet
demonstrating superior performance. The findings underscore
the importance of robust training regimes in achieving accurate
and efficient EEG classification, providing valuable insights for
deploying AI models in clinical practice.
\end{abstract}


\section{Introduction}

The accurate interpretation of electroencephalogram (EEG) signals are essential for diagnosing neurological conditions such as seizures, Generalized Periodic Discharges (GPD), Lateralized Periodic Discharges (LPD), and other harmful brain activities \cite{b1}. Traditionally, expert neurologists manually review these recordings—a process that is time-consuming, subjective, and difficult to scale \cite{b2}. The increasing availability of large EEG datasets, coupled with advancements in machine learning \cite{b3}, offers the potential to automate this process, enabling the efficient analysis of complex brain activity patterns \cite{b4}.

EEG signals are non-stationary and present complex time-frequency characteristics, making them challenging to interpret using traditional methods. Transforming EEG data into time-frequency representations, such as spectrograms generated through the Continuous Wavelet Transform (CWT) \cite{b5}, enables machine learning models to capture the dynamic oscillatory patterns of neural activity \cite{b6}. Spectrograms provide a structured visual format ideal for models to process, especially in deep learning contexts \cite{b7,b8}.

Recent advances in deep learning have positioned Convolutional Neural Networks (CNNs) \cite{b9} and Vision Transformers (ViTs) \cite{b10} as promising approaches for EEG classification \cite{b8}. CNNs excel at extracting spatial patterns from spectrograms \cite{b11}, while Vision Transformers, with their self-attention mechanisms, capture long-range dependencies across spectrogram data \cite{b13,b6}. Additionally, specialized architectures like EEGNet, designed specifically for EEG data, offer efficient and compact solutions for EEG-based tasks \cite{b12}.

This study focuses on modelling how expert neurologists classify EEG signals—whether waveforms or spectrograms—into categories of harmful brain activity. Expert labels are generated based on their consensus votes over EEG data, and the aim is to replicate their decision-making process through machine learning models \cite{b2,b4}. The study includes both deep learning models and Gradient Boosting Models (GBMs) such as CatBoost, despite the need for feature engineering with GBMs in unstructured data like EEG \cite{b14}. GBMs are included to provide a baseline for comparison, emphasizing interpretability and computational efficiency \cite{b1}, which are crucial in clinical settings \cite{b13}.

A key insight from recent research and experimental platforms (e.g., the work from Barnett et al. \cite{b4} and public experimentation at the 2024 HMS: Harmful Brain Activity Classification Contest hosted by Kaggle) is that training strategies—such as data preprocessing, augmentation, and multi-stage training—often plays a larger role in performance than the specific architecture itself \cite{b2,b6}. By comparing CNNs, ViTs, and CatBoost under uniform training strategies, this study identifies the factors that most significantly influence classification accuracy in EEG data. This research builds on the insights gained from previous works and experimental results, aiming to identify the most effective strategies for deploying EEG analysis systems in clinical settings.

\section{Literature Survey}
This literature survey reviews key developments in machine learning models, particularly deep learning, for classifying harmful brain activities from EEG data. It covers Spectrogram-based EEG analysis, transformer-based architectures, and how training strategies affect model performance.

\subsection{Machine Learning in EEG Classification} EEG signals, known for their non-stationary and noisy nature, have long posed challenges in effective classification. Early approaches by Shoeb et al. \cite{b1} demonstrated the potential of machine learning for automating manual EEG review, particularly using Support Vector Machines (SVMs) for seizure detection. This foundational work set the stage for the adoption of more advanced techniques like deep learning \cite{b15}.

\subsection{Deep Learning Architectures in EEG} Recent advancements in deep learning, specifically the use of Convolutional Neural Networks (CNNs) and Recurrent Neural Networks (RNNs), have reshaped EEG analysis. Roy et al. \cite{b6} and Acharya et al. \cite{b9} demonstrated the effectiveness of CNNs in capturing time-frequency representations of EEG data, improving the detection of seizures and other harmful brain activities. Lawhern et al. \cite{b12} introduced EEGNet, a compact neural network optimized for EEG tasks, providing a lightweight yet effective solution for EEG classification, especially in clinical environments with limited computational resources.

\subsection{Spectrograms as Preprocessing for EEG Analysis} Spectrograms have become a common preprocessing step, transforming EEG signals into time-frequency representations \cite{b16}. Bashivan et al. \cite{b5} and Roy et al. \cite{b6} demonstrated that using spectrograms significantly enhances model performance by enabling models to capture both temporal and frequency-based features. These representations are particularly useful for deep learning models like CNNs and transformers, which excel in image-based and sequential data tasks \cite{b17,b15}.

\subsection{Transformer Based Models} Transformers \cite{b19} have emerged as a powerful alternative to traditional deep learning methods in EEG classification, especially for handling complex data. Vision Transformers (ViTs), introduced by Dosovitskiy et al. \cite{b10}, leverage self-attention mechanisms to model long-range dependencies, offering improved performance on EEG data. Ortega Caro et al. \cite{b18} demonstrated the utility of transformers in fMRI and brain activity recordings with BrainLM, and Kerr et al. \cite{b15} showed how transformer-based models offer strong potential for continuous seizure detection and brain activity classification.

\subsection{Comparative Analysis of Machine Learning Models} Several studies have conducted comparative analyses of traditional machine learning models versus deep learning architectures for EEG classification. Schirrmeister et al. \cite{b11} compared SVMs, Random Forests, and deep learning models (CNNs and LSTMs), finding that deep learning approaches consistently outperformed traditional methods. Pfeffer et al. \cite{b13} showed that transformers \cite{b18}, although more computationally demanding, achieved superior performance over CNNs in handling large, complex EEG datasets.

\subsection{Practical Adoption Challenges} Despite advancements, challenges remain in the clinical adoption of deep learning models for EEG classification. Shoeb \cite{b1} and Acharya et al. \cite{b13} highlighted the difficulties in scaling models from controlled, small datasets to real-world clinical environments. Additionally, interpretability and computational resource requirements remain significant barriers to the deployment of deep learning models in practice, as discussed by Schirrmeister et al. \cite{b11} and Jing et al. \cite{b2}.

\subsection{Recent Advances and Training Techniques} Recent studies have focused on improving the performance and efficiency of deep learning models for EEG analysis. Kerr et al. \cite{b15} explored transformer-based models for continuous seizure detection, highlighting their ability to reduce false positives while maintaining high sensitivity. Additionally, public experimentation platforms like HMS: HBAC 2024 have provided opportunities to rapidly test a diverse variety of methods, suggesting that training protocols often outweigh architectural choices in their impact on performance \cite{b6}.
\section{Methodology}

\subsection{Data Characteristics and Representation}
The training dataset contains 106,800 rows with 17,089 unique EEG IDs, 11,138 unique spectrogram IDs, and 1,950 unique patients. Each row represents a specific 50-second EEG window and a corresponding 600-second spectrogram window. The models predict the event occurring in the middle 10 seconds of these windows. EEG data is stored in parquet files, each containing multiple rows, while spectrogram data is stored in fewer parquet files due to shared time windows across rows.

\begin{figure}[h!]
    \centering
    \includegraphics[width=1\linewidth]{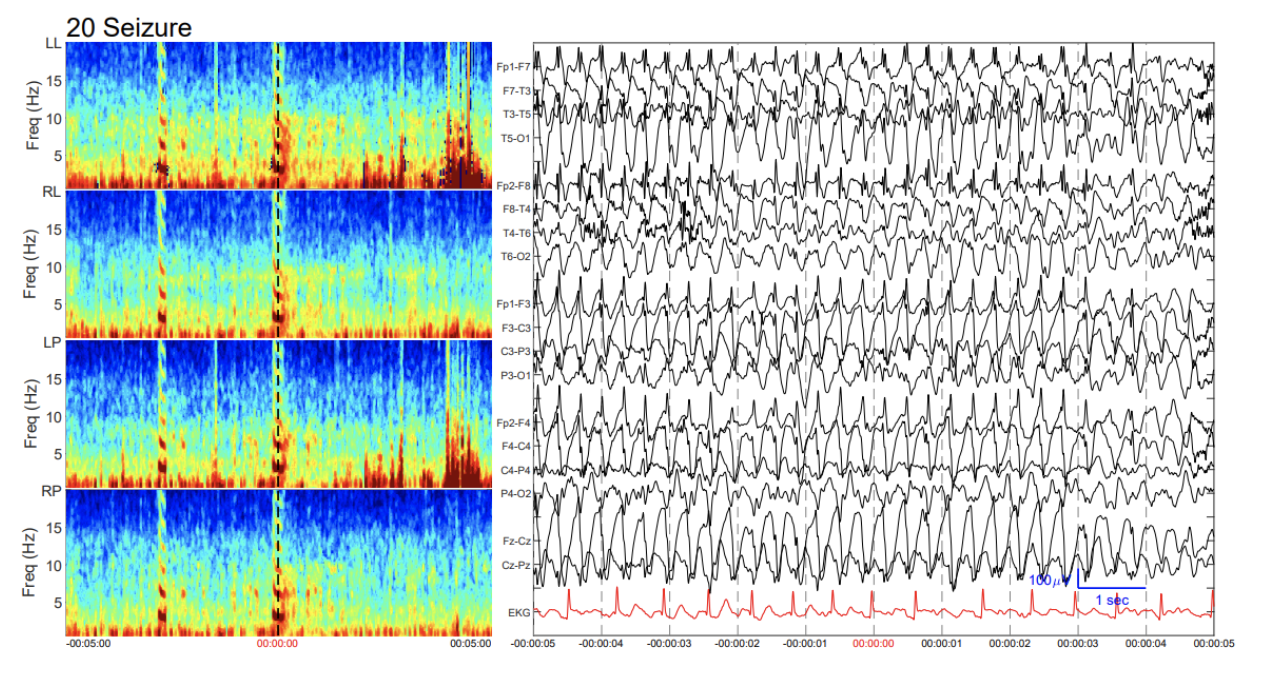}
    \caption{EEG Waveform and Corresponding Spectrogram from the Dataset}
    \label{fig:enter-label}
\end{figure}

\begin{table}[h!]
\centering
\scriptsize
\setlength{\tabcolsep}{2pt} 
\renewcommand{\arraystretch}{1.2} 

\vspace{10pt} 
\begin{tabular}{|p{0.10\linewidth}|p{0.08\linewidth}|p{0.15\linewidth}|p{0.08\linewidth}|p{0.08\linewidth}|p{0.08\linewidth}|p{0.08\linewidth}|p{0.08\linewidth}|p{0.08\linewidth}|}
\hline
\textbf{EEG ID} & \textbf{Patient ID} & \textbf{Expert Consensus} & \textbf{Seizure Vote} & \textbf{LPD Vote} & \textbf{GPD Vote} & \textbf{LRDA Vote} & \textbf{GRDA Vote} & \textbf{Other Vote} \\
\hline
16 & 42516 & Seizure & 3 & 0 & 0 & 0 & 0 & 0 \\
\hline
22 & 30539 & GPD & 0 & 0 & 5 & 0 & 1 & 5 \\
\hline
\end{tabular}
\vspace{10pt} 
\caption{EEG Data Votes}
\end{table}

The Double Banana Montage \cite{b20} was applied to generate differential signals across four key brain regions: Left Temporal, Left Parasagittal, Right Temporal, and Right Parasagittal. The montage reduces common-mode noise and emphasizes spatial distinctions, making it suitable for spectrogram generation through the Continuous Wavelet Transform (CWT).

\begin{scriptsize} 
\begin{equation}
\text{LL Spec} = \frac{1}{4} \left( \text{spec(Fp1 - F7)} + \text{spec(F7 - T3)} + \text{spec(T3 - T5)} + \text{spec(T5 - O1)} \right)
\label{eq:ll_spec}
\end{equation}
\end{scriptsize}

Equation \ref{eq:ll_spec} represents the process of generating the LL Spectrogram by averaging the spectrograms computed from the differential signals of adjacent electrodes in the Left Temporal Chain, using the Continuous Wavelet Transform (CWT) function for time-frequency representation.

\subsection{Data Preprocessing}
\begin{itemize}
    \item EEG-HR: High-resolution EEG waveform segments, 50 seconds long, sampled at 200 Hz.
    \item Spec-HR: High-resolution spectrograms generated from the raw EEG Data, using CWT and the Double Banana montage, focusing on a frequency range of 0.5–40 Hz, 40 scales, and a stride of 16.
    \item Spec-LR: Low-resolution spectrograms generated from longer 10-minute EEG segments, providing broader but less temporally granular insights. \cite{b2}
\end{itemize}

A Butterworth bandpass filter (0.5–20 Hz) was applied with clipping to ensure data consistency across electrodes.

\begin{figure}[h!]
    \centering
    \includegraphics[width=1\linewidth]{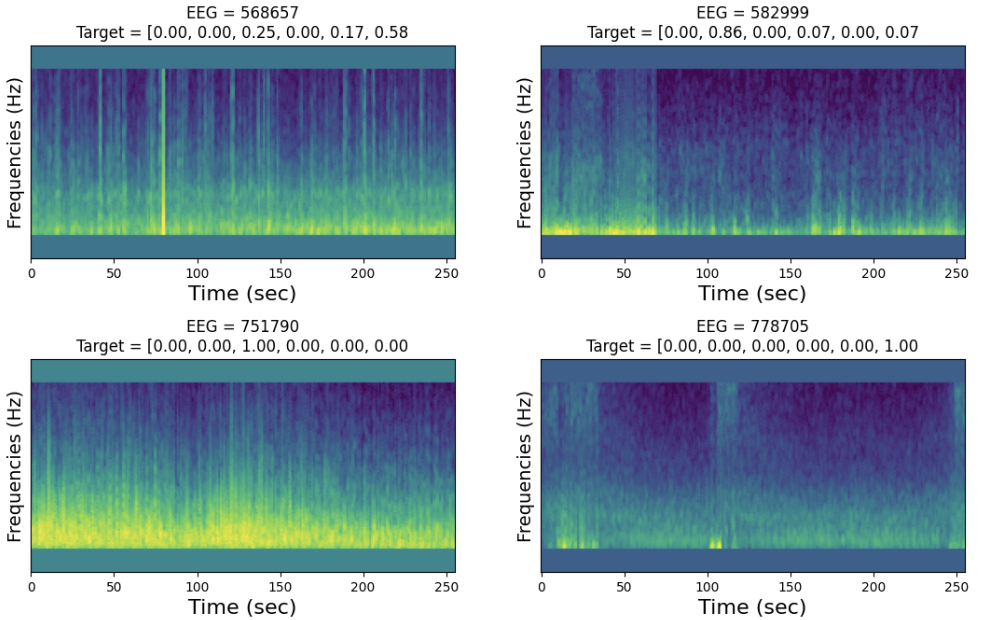}
    \caption{Spectrograms Generated from Preprocessed EEG Data}
    \label{fig:enter-label}
\end{figure}

\subsection{Data Augmentation}

For 2-D models, several augmentation techniques were employed. \textbf{XY Masking} involved random masking of 1 to 8 spectrogram nodes in 50\% of the samples, simulating missing or noisy data. \textbf{Mixup} combined spectrogram images and labels during training, promoting generalization by generating new data points as weighted combinations of the original samples. \textbf{Window Shifting} augmented the training data by shifting the central 50s window by up to 5 seconds, introducing temporal variability in the data.

For 1-D models, augmentations such as the \textbf{Left-Right Flip} were applied, randomly flipping the EEG signal temporally to simulate different orientations of brain activity. Additionally, the \textbf{Brain-Side Flip} randomly switched the left-right brain data, further capturing variability in spatial brain signal distributions.

These augmentations helped capture spatial and temporal variability in the brain's electrical activity, improving the model’s ability to generalize to unseen EEG data.

\subsection{Training Procedure}
The training procedure follows a two-stage approach:

Stage 1: High-Confidence Data:  
Models were trained using data with $\geq$ 10 expert votes, focusing on reliable and consensus-labeled examples. Cosine Annealing's learning rate scheduler with an initial learning rate of 0.0012 and a batch size of 32 was used for 5 epochs.

\begin{figure}[h!]
    \centering
    \includegraphics[width=0.8\linewidth]{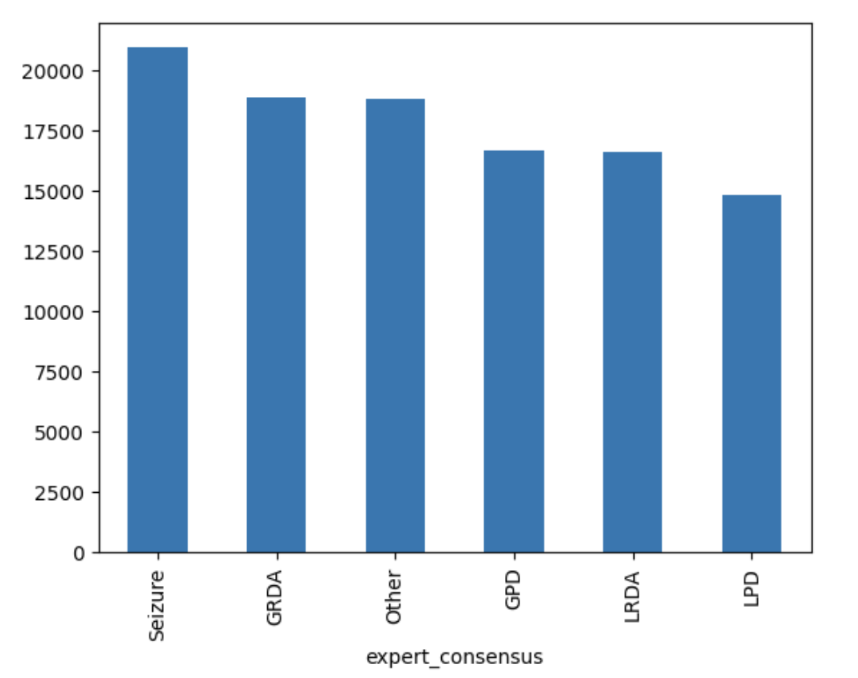}
    \caption{Expert Consensus Classes Distribution}
    \label{fig:enter-label}
\end{figure}

Stage 2: Full Dataset:  
The models were then fine-tuned using the entire dataset, including low-vote samples, with progressively reduced weight for lower-vote data during training. Group K-Fold cross-validation was used with a 5-fold split, ensuring patient-specific samples were not split between training and validation sets.

\subsection{Model Architectures}

\textbf{MLP Multilabel Classifier:} Baseline model.

\textbf{WaveNet:} WaveNet is adapted for raw EEG waveform processing, learning long-term dependencies through causal convolutions \cite{b15}.

\textbf{CatBoost:} CatBoost is a gradient-boosting model used for both spectrogram and waveform features, offering interpretability and efficiency \cite{b10, b13}.

\textbf{EfficientNetB2:} EfficientNetB2 is pre-trained on ImageNet and processes spectrograms using its CNN architecture \cite{b6}.

\textbf{EEGNet:} EEGNet is a compact neural network optimized for EEG tasks, trained directly on raw EEG features \cite{b1}.

\textbf{ResNet34d:} CNN pre-trained on the CIFAR-10 dataset, with residual connections to capture hierarchical patterns \cite{b5}.

\textbf{EEGNet \& EfficientNet/TinyViT Ensemble:} This setup combines a 1-D CNN model (EEGNet) with either EfficientNet or TinyViT as the 2-D backbone to capture temporal dynamics from EEG waveforms and feature embeddings from spectrograms. A two-stage training process is employed, where each ensemble is first trained on high-confidence samples for 5 epochs and then fine-tuned on the broader dataset for an additional 15 epochs, for a total of 20 epochs. The outputs from the 1-D EEGNet and the 2-D EfficientNet or TinyViT models are concatenated in a final layer, allowing for a direct comparison between EfficientNet and TinyViT in terms of their ability to process high-resolution spectrograms for EEG classification \cite{b5, b18}.

\begin{figure}[h!]
    \centering
    \includegraphics[width=0.5\linewidth]{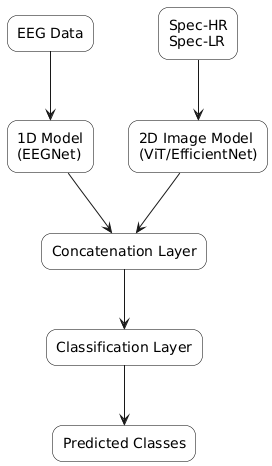}
    \caption{Multi-Modal EEG Classification Model Architecture }
    \label{fig:multi_modal}
\end{figure}

\subsection{Model Evaluation}
The performance of the models was evaluated using KL Divergence across cross-validation folds, assessing the alignment between model predictions and the true label distributions \cite{b11}.

\begin{equation}
D_{\text{KL}}(P \parallel Q) = \sum_{i} P(i) \log \frac{P(i)}{Q(i)}
\label{eq:kl_divergence}
\end{equation}

As shown in Equation~\ref{eq:kl_divergence}, the Kullback-Leibler (KL) Divergence measures the difference between two probability distributions $P$ and $Q$.

\vspace{10pt} 

\section{Results}

This section provides a comprehensive analysis of the models used in the study, evaluating their performance and discussing key insights gained from training and testing on the EEG dataset. 

The selected epoch count aimed to balance computational efficiency with effective model convergence. For smaller models, we limited training to 4–5 epochs, as extending further led to overfitting and did not yield significant gains in performance. This conservative approach prevented excessive training on limited data, avoiding diminishing returns. Although preliminary tests indicated that additional epochs had limited impact, future exploration with increased epoch counts could provide further insights into model robustness and convergence behaviour.

\begin{table}[h!]
\centering
\scriptsize
\setlength{\tabcolsep}{3pt} 
\renewcommand{\arraystretch}{1.2} 

\vspace{10pt} 
\begin{tabular}{|p{0.22\linewidth}|p{0.20\linewidth}|p{0.12\linewidth}|p{0.12\linewidth}|p{0.10\linewidth}|}
\hline
\textbf{Models} & \textbf{Dataset Used} & \textbf{CV Score} &  \textbf{Infer Score} & \textbf{Epochs} \\
\hline
MLP multilabel classifier & Spec-LR & 1.037 & 0.77 & 5  \\
\hline
WaveNet & EEG-HR & 0.91 & 0.66 & 5  \\
\hline
Catboost & Spec-LR + Spec-HR & 0.74 & 0.60 & 5  \\
\hline
EfficientNetB2 & Spec-LR & 0.72 & 0.57 & 5  \\
\hline
EEGNet & EEG-HR & 0.69 & 0.58 & 5  \\
\hline
ResNet34d & Spec-LR & 0.70 & 0.49 & 5  \\
\hline
EfficientNetB0 & Spec-LR + Spec-HR & 0.59 & 0.44 & 4 \\
\hline
Multi Stage with EfficientNet Model & EEG-HR + Spec-LR + Spec-HR & 0.34 & 0.26 & 20  \\
\hline
Multi Stage with TinyViT & EEG-HR + Spec-LR + Spec-HR & 0.24 & 0.24 & 20  \\
\hline
\end{tabular}
\vspace{10pt} 
\caption{Model Performance Comparison}
\end{table}

\subsection{Model Performance Analysis}

\textbf{MLP Multilabel Classifier:} The MLP, a shallow architecture lacking the ability to capture complex spatial and temporal patterns in the EEG data, performed poorly, with a cross-validation (CV) score of 1.037 and an inference score of 0.77. This highlights the limitations of simple architectures when dealing with intricate EEG data, which requires advanced feature extraction methods for accurate classification.

\textbf{WaveNet:} WaveNet, designed for sequential data, was applied to raw EEG waveforms and performed moderately well, achieving a CV score of 0.91 and an inference score of 0.66. The model’s ability to capture long-term dependencies in raw EEG data through causal convolutions was beneficial, though it struggled with spectrogram data, which requires a different processing approach.

\textbf{CatBoost:} CatBoost, a gradient-boosting model applied to both raw EEG and spectrogram data, showed reasonable results with a CV score of 0.74 and an inference score of 0.60. Although it excels in handling structured data, its lack of deep spatial and temporal feature extraction capabilities limited its overall performance on the unstructured, high-resolution EEG spectrogram data.

\textbf{EfficientNetB2 \& EfficientNetB0:} EfficientNetB2 performed better than its smaller variant EfficientNetB0, with a CV score of 0.72 and an inference score of 0.57. EfficientNetB0, after undergoing multi-stage training (with high-confidence labels followed by full dataset fine-tuning), improved significantly, achieving a CV score of 0.34 and an inference score of 0.26. This multi-stage training approach helped mitigate overfitting and align the model better with the expert-labeled test data.

\begin{figure}[h!]
    \centering
    \includegraphics[width=0.8\linewidth]{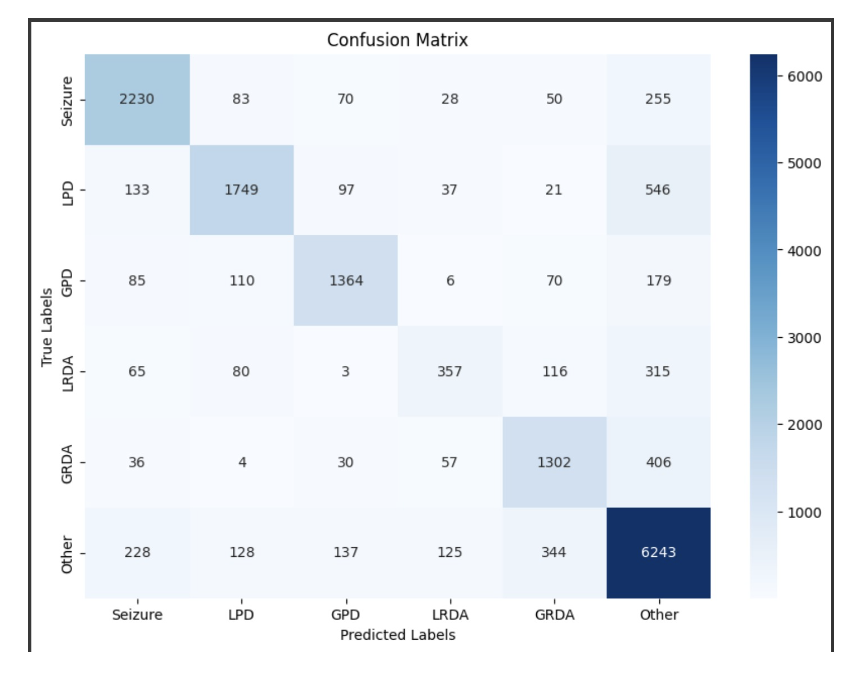}
    \caption{Confusion Matrix for EfficientNetB2 }
    \label{fig:confusion_matrix}
\end{figure}

\textbf{Analysis of the Confusion Matrix: }Although confusion matrices provide a breakdown of true positives, false positives, and misclassifications across different classes, they offer limited value for EEG classification tasks like ours, where understanding the model’s probabilistic predictions is more critical. The confusion matrix presented (see Figure~\ref{fig:confusion_matrix}) highlights the model’s ability to classify harmful brain activities, but it oversimplifies the complex decision-making process involved in EEG analysis. 

Our models produce probability distributions across multiple potential classes, and focusing solely on the final predicted labels, as in a confusion matrix, neglects the importance of prediction confidence. In clinical applications, this confidence is vital, particularly when distinguishing between closely related classes like Seizures and LPDs. Therefore, metrics such as Kullback-Leibler (KL) divergence provides a more nuanced evaluation by assessing how well the predicted probability distribution aligns with the true class distribution, reflecting the model’s uncertainty and confidence more effectively than traditional evaluation methods.

\textbf{EEGNet:} Specifically designed for EEG data, EEGNet demonstrated strong performance on raw EEG data, with a CV score of 0.69 and an inference score of 0.58. Its architecture, which captures spatial-temporal EEG patterns without requiring a frequency transformation, allowed it to handle raw data efficiently. However, it was slightly outperformed by models trained on spectrogram data, which provided richer feature representations.

\textbf{ResNet34d:} ResNet34d, used for spectrogram analysis, performed well on low-resolution spectrograms, with a CV score of 0.715 and an inference score of 0.49. The residual connections in ResNet allowed the model to capture hierarchical patterns in the spectrogram data. However, its depth was not sufficient to outperform more modern transformer-based approaches, particularly in the multi-stage setups.

\textbf{TinyViT and Multi-Stage EfficientNet Ensemble:} TinyViT, a variant of the Vision Transformer (ViT), was selected for its compact and efficient architecture, particularly suited to noisy EEG data. The multi-stage TinyViT model achieved the best results, with a CV score of 0.24 and an inference score of 0.24. EfficientNetB0, using a similar multi-stage approach, also improved significantly (CV score: 0.34, inference score: 0.26). However, TinyViT outperformed EfficientNetB0, particularly in terms of generalization and computational efficiency, due to its pretraining and knowledge distillation techniques.

\subsection{Key Insights and Discussion}

\textbf{Training Strategy Over Model Choice:} Across all models, the two-stage training approach—pre-training on high-confidence samples followed by fine-tuning the broader dataset—proved to be the most effective strategy. This finding aligns with existing research, such as SPaRCNet and Improving Clinician Performance, which emphasizes the importance of training strategies in handling variable and noisy EEG data. TinyViT and EfficientNet models, when trained in this staged manner demonstrated significant improvements over their single-stage counterparts, supporting the idea that a tailored training regimen is often more crucial than the model architecture itself.

\textbf{Preprocessing and Data Representation:} Spectrogram-based models consistently outperformed those trained solely on raw EEG data. The Continuous Wavelet Transform (CWT), used to generate time-frequency spectrograms, allowed models to capture complex oscillatory patterns that were not easily extractable from raw waveforms. The Double Banana Montage further enhanced feature extraction by reducing noise and emphasizing spatial information, much like expert neurologists’ clinical practice of examining different brain regions through montages.

\textbf{Augmentation and Regularization:} Augmentation techniques, particularly those mimicking real-world variability (e.g., channel flipping, signal distortion), played a critical role in reducing overfitting. Models trained with such augmentations, including TinyViT and EfficientNet, showed better generalization during cross-validation and inference. These augmentations helped simulate clinical scenarios where EEG data might be noisier or more difficult to interpret.

\textbf{Multi-Modal Architectures:} Models that leveraged both raw EEG waveforms and spectrograms, such as the multi-stage TinyViT and EfficientNetB0, benefited from the ability to extract complementary features from both time-domain and frequency-domain representations. This multi-modal approach mirrors clinical practices, where EEG readings are evaluated from multiple perspectives (e.g., time-domain waveforms and frequency-domain analyses) to provide a comprehensive view of brain activity.

\textbf{State-of-the-Art Models: Not Always the Best Choice:} Surprisingly, smaller and more efficient models like TinyViT outperformed larger, state-of-the-art architectures such as the full-size Vision Transformers (ViT). While larger models often overfit the small and noisy EEG dataset, TinyViT’s compact architecture, when paired with pretraining and multi-stage training, provided more consistent and robust performance. This underscores the importance of choosing models that balance complexity with the specific characteristics of the dataset at hand.

\textbf{Limitations:} This study is limited by dataset size, class imbalance, and dependence on high-quality spectrograms, which may impact generalizability to more diverse and minimally processed EEG datasets. Additionally, our approach does not fully explore model ensembling, which could enhance robustness by integrating complementary strengths across architectures. Further work is needed to address interpretability challenges and to develop preprocessing techniques that effectively expand the dataset while preserving clinical relevance.

\section{Conclusion}

This study explored a variety of machine learning architectures and training strategies for EEG classification, demonstrating that the most significant performance improvements stem not solely from adopting state-of-the-art models, but from employing an optimized combination of data preprocessing, multi-stage training strategies, and effective augmentation techniques. The results show that models like TinyViT and EfficientNet, when paired with multi-stage training and multimodal data representations, significantly outperform their single-stage counterparts. This highlights the critical role that training regimes and preprocessing play in handling the inherent noise and complexity of EEG data.

The insights gained from comparing these models under uniform training conditions build upon and extend the work of previous research, such as SPaRCNet \cite{b2} and Improving Clinician Performance in EEG Classification \cite{b4}, by demonstrating the potential of AI-based classification tools to assist clinicians in diagnosing harmful brain activities. The findings underscore the importance of tailoring models to the specific characteristics of EEG signals, focusing on strategies that enhance both model robustness and generalization to clinical datasets. As the field advances, future research should prioritize refining these techniques and further exploring multimodal data integration to bolster the clinical applicability and reliability of AI-driven EEG analysis.


\begin{thebibliography}{00}

\bibitem{b1} A. H. Shoeb, ``Application of machine learning to epileptic seizure onset detection and treatment'', Doctoral Dissertation, Massachusetts Institute of Technology, 2009.

\bibitem{b2} J. Jing, W. Ge, S. Hong, M. B. Fernandes, Z. Lin, C. Yang, ... \& M. B. Westover, ``Development of expert-level classification of seizures and rhythmic and periodic patterns during EEG interpretation'', \textit{Neurology}, vol. 100, no. 17, pp. e1750-e1762, 2023.

\bibitem{b3} G. Amrani, A. Adadi, M. Berrada, Z. Souirti, \& S. Boujraf, ``EEG signal analysis using deep learning: A systematic literature review'', \textit{In 2021 Fifth International Conference On Intelligent Computing in Data Sciences (ICDS)}, pp. 1-8, IEEE, 2021.

\bibitem{b4} A. J. Barnett, Z. Guo, J. Jing, W. Ge, P. W. Kaplan, W. Y. Kong, ... \& M. B. Westover, ``Improving Clinician Performance in Classifying EEG Patterns on the Ictal–Interictal Injury Continuum Using Interpretable Machine Learning'', \textit{NEJM AI}, vol. 1, no. 6, AIoa2300331, 2024.

\bibitem{b5} P. Bashivan, I. Rish, M. Yeasin, \& N. C. Codella, ``Learning representations from EEG with deep recurrent-convolutional neural networks'', \textit{CoRR}, abs/1511.06448, 2015.

\bibitem{b6} Y. Roy, H. Banville, I. Albuquerque, A. Gramfort, T. H. Falk, \& J. Faubert, ``Deep learning-based electroencephalography analysis: A systematic review'', \textit{Journal of Neural Engineering}, vol. 16, no. 5, 051001, 2019.

\bibitem{b7} A. Chaddad, Y. Wu, R. Kateb, \& A. Bouridane, ``Electroencephalography Signal Processing: A comprehensive review and analysis of methods and techniques'', \textit{Sensors}, vol. 23, no. 14, 6434, 2023.

\bibitem{b8} Y. Sun, W. Jin, X. Si, X. Zhang, J. Cao, L. Wang, S. Yin, \& D. Ming, ``Continuous Seizure Detection Based on Transformer and Long-Term iEEG'', \textit{IEEE Journal of Biomedical and Health Informatics}, vol. 26, pp. 5418-5427, 2022.

\bibitem{b9} U. R. Acharya, S. L. Oh, Y. Hagiwara, J. H. Tan, \& H. Adeli, ``Deep convolutional neural network for the automated detection and diagnosis of seizure using EEG signals'', \textit{Computers in Biology and Medicine}, vol. 100, pp. 270-278, 2018.

\bibitem{b10} A. Dosovitskiy, L. Beyer, A. Kolesnikov, D. Weissenborn, X. Zhai, T. Unterthiner, M. Dehghani, M. Minderer, G. Heigold, S. Gelly, J. Uszkoreit, \& N. Houlsby, ``An Image is Worth 16x16 Words: Transformers for Image Recognition at Scale'', \textit{ArXiv}, abs/2010.11929, 2020.

\bibitem{b11} R. T. Schirrmeister, J. T. Springenberg, L. D. J. Fiederer, M. Glasstetter, K. Eggensperger, M. Tangermann, F. Hutter, W. Burgard, \& T. Ball, ``Deep learning with convolutional neural networks for EEG decoding and visualization'', \textit{Human Brain Mapping}, vol. 38, no. 11, pp. 5391-5420, 2017.

\bibitem{b12} V. J. Lawhern, A. J. Solon, N. R. Waytowich, S. M. Gordon, C. P. Hung, \& B. J. Lance, ``EEGNet: A compact convolutional neural network for EEG-based brain–computer interfaces'', \textit{Journal of Neural Engineering}, vol. 15, no. 5, 056013, 2018.

\bibitem{b13} M. A. Pfeffer, S. S. H. Ling, \& J. K. W. Wong, ``Exploring the Frontier: Transformer-Based Models in EEG Signal Analysis for Brain-Computer Interfaces'', \textit{Computers in Biology and Medicine}, 108705, 2024.

\bibitem{b14} M. Saeidi, W. Karwowski, F. V. Farahani, K. Fiok, R. Taiar, P. A. Hancock, \& A. Al-Juaid, ``Neural decoding of EEG signals with machine learning: A systematic review'', \textit{Brain Sciences}, vol. 11, no. 11, 1525, 2021.

\bibitem{b15} W. T. Kerr, K. N. McFarlane, \& G. F. Pucci, ``The present and future of seizure detection, prediction, and forecasting with machine learning'', \textit{Frontiers in Neurology}, vol. 15, 1425490, 2024.

\bibitem{b16} D. P. Subha, P. K. Joseph, U. R. Acharya, \& C. M. Lim, ``EEG signal analysis: A survey'', \textit{Journal of Medical Systems}, vol. 34, no. 2, pp. 195-212, 2010.

\bibitem{b17} B. Abibullaev, A. Keutayeva, \& A. Zollanvari, ``Deep Learning in EEG-Based BCIs: A Comprehensive Review of Transformer Models, Advantages, Challenges, and Applications'', \textit{IEEE Access}, vol. 11, pp. 127271-127301, 2023.

\bibitem{b18} J. Ortega Caro, A. H. Oliveira Fonseca, C. Averill, S. A. Rizvi, M. Rosati, J. L. Cross, ... \& D. van Dijk, ``BrainLM: A foundation model for brain activity recordings'', \textit{bioRxiv}, 2023-09.

\bibitem{b19} A. Vaswani, N. Shazeer, N. Parmar, J. Uszkoreit, L. Jones, A. N. Gomez, L. Kaiser, \& I. Polosukhin, ``Attention is all you need'', \textit{Advances in Neural Information Processing Systems}, 2017.

\bibitem{b20} J. Ochoa, W. Gonzalez, R. Bautista, \& J. DeCerce, ``32-Channel banana-avg montage is better than 16-channel double banana montage to detect epileptiform discharges in routine EEGs'', \textit{Clinical Neurophysiology}, vol. 119, no. 10, pp. 2185-2187, 2008.

\bibitem{b21} Y. R. Tabar, \& U. Halici, ``A novel deep learning approach for classification of EEG motor imagery signals'', \textit{Journal of Neural Engineering}, vol. 14, no. 1, 016003, 2017.

\bibitem{b22} N. Padfield, J. Zabalza, H. Zhao, V. Masero, \& J. Ren, ``EEG-based brain-computer interfaces using motor-imagery: Techniques and challenges'', \textit{Sensors}, vol. 19, no. 6, 1423, 2019.

\end{thebibliography}
\end{document}